\documentclass[10pt,twocolumn,letterpaper]{article}

\usepackage{cvpr}
\usepackage{times}
\usepackage{epsfig}
\usepackage{graphicx}
\usepackage{gensymb} % For the \degree symbol
\usepackage{color}
\usepackage{amsmath,amsfonts,amssymb,mathrsfs}
\usepackage{graphicx}
\usepackage{amsmath}
\usepackage{amssymb}
\usepackage{comment}
\usepackage{epstopdf}
\usepackage{tabularx}
\usepackage{bbold}
\usepackage{eqnarray}
\usepackage{algorithm}
\usepackage{array,booktabs}
\usepackage{enumitem}

\definecolor{myGray}{rgb}{0.6,0.6,0.6}

\DeclareMathOperator*{\rot}{rotate2D}

\DeclareMathOperator*{\stddev}{std}

\newcommand{\px}{\,\textit{px}\;}
\newcommand{\shortTitle}[1]{\noindent{\normalfont\normalsize\bfseries{#1}}\hspace{6pt}}

\newcommand{\gt}[1]{\textcolor{myGray}{#1}}

\def\ie{\emph{i.e.}~}
\def\eg{\emph{e.g.}~}

\def\etal{\emph{et al.}}
\newcommand{\fig}{Fig.~}
\newcommand{\eq}{Eq.\,}
\newcommand{\sect}{Section~}

\newcommand{\m}{\nobreak\hspace{-.2em}}
\newcommand{\n}{\nobreak\hspace{-.22em}}
\newcommand{\mm}{\nobreak\hspace{-.4em}}
\newcommand{\nc}{{N.C.}}
\newcommand{\degreeSymbol}{$^{\circ}$}

\renewcommand{\ll}{{\ell\hspace{-.1em}+\hspace{-.1em}1}}
\renewcommand{\lll}{{\ell}}
\newcommand{\conv}{\ast{}_{_{\hspace{-.15em}2D}}}
\newcommand{\convv}{\ast}

\newcommand{\gaussKernels}{0}
\newcommand{\motionKernels}{1}
\newcommand{\poolingKernels}{2}
\newcommand{\decodingKernels}{3}
\newcommand{\outputKernels}{4}

\newcommand{\thetaIn}{\theta^{\lll}}
\newcommand{\thetaOut}{\theta^{\ll}}

\newcommand{\hangBox}[1]{%^
  \begin{minipage}[t]{\textwidth}% Top-hanging minipage, will align on
  \begin{tabbing} % tabbing so that minipage shrinks to fit
  ~\\[-\baselineskip] % Make first line zero-height
  #1 % Include user's text
  \end{tabbing}%^
  \end{minipage}}

\newcommand{\middleburyRow}[1]{
  \hangBox{\includegraphics[width=20mm]{Middlebury-img-#1.jpg}}&
  \hangBox{\includegraphics[width=20mm]{Middlebury-gt-#1.jpg}}&
  \hangBox{\includegraphics[width=20mm]{Middlebury-us-#1.jpg}}&
  \hangBox{\includegraphics[width=20mm]{Middlebury-flownets-#1.jpg}}\\
}

\newcommand{\sintelRow}[1]{
  \hangBox{\includegraphics[width=20mm]{Sintel-img-#1.png}}&
  \hangBox{\includegraphics[width=20mm]{Sintel-gt-#1.png}}&
  \hangBox{\includegraphics[width=20mm]{Sintel-us-#1.png}}&
  \hangBox{\includegraphics[width=20mm]{Sintel-flownets-#1.png}}\\
}

\newcommand{\sintelRowSimple}[1]{
  \hangBox{\includegraphics[width=20mm]{Sintel2-#1.png}}&
  \hangBox{\includegraphics[width=20mm]{Sintel2-#1gt.png}}&
  \hangBox{\includegraphics[width=20mm]{Sintel2-#1us.png}}&
  \hangBox{\includegraphics[width=20mm]{Sintel2-#1occ.png}}\\
}

\graphicspath{
  {../Figures}
}

\cvprfinalcopy

\def\cvprPaperID{216}

\begin{document}

\title{\vspace{-6pt}Learning to Extract Motion from Videos in Convolutional Neural Networks\vspace{-5pt}}

\author{
  Damien Teney ~~~~~~~~~~~~~Martial Hebert
  \vspace{4pt}\\
  %{\tt\small dteney@andrew.cmu.edu~~hebert@ri.cmu.edu}
  %\vspace{-2pt}\\
  The Robotics Institute\\Carnegie Mellon University\\Pittsburgh, PA, USA
  \vspace{-4pt}
}

\maketitle

\thispagestyle{empty} % No number on first page
\pagestyle{empty} % No number on subsequent pages

\begin{abstract}
This paper shows how to extract dense optical flow from videos with a convolutional neural network (CNN). The proposed model constitutes a potential building block for deeper architectures to allow using motion without resorting to an external algorithm, \eg for recognition in videos. We derive our network architecture from signal processing principles to provide desired invariances to image contrast, phase and texture. We constrain weights within the network to enforce strict rotation invariance and substantially reduce the number of parameters to learn. We demonstrate end-to-end training on only 8 sequences of the Middlebury dataset, orders of magnitude less than competing CNN-based motion estimation methods, and obtain comparable performance to classical methods on the Middlebury benchmark. Importantly, our method outputs a distributed representation of motion that allows representing multiple, transparent motions, and dynamic textures. Our contributions on network design and rotation invariance offer insights nonspecific to motion estimation.
\end{abstract}

%potentially applicable beyond 
%This judicious design leads to a shallow network trainable on little data with no need for artificial augmentations. 
%On the Sintel benchmark, We obtain reasonable performance, but show limitations in handling large displacements. 
%that go beyond traditional flow methods. It can capture a much wider range of phenomena such as

%====================================================================================================================
\section{Introduction}

The success of convolutional neural networks (CNNs) on image-based tasks, from object recognition to semantic segmentation or geometry prediction, has inspired similar developments with videos. Example applications include activity recognition \cite{simonyan2014,wu2015,ye2015}, scene classification \cite{tran2014}, or semantic segmentation of scenes with dynamic textures \cite{teney2014}. The appeal of CNNs is to be trainable end-to-end, \ie taking raw pixel values as input, and learning their mapping to the output of choice, identifying appropriate intermediate representations in the layers of the network. The natural application of this paradigm to videos involves a 3D volume of pixels as input, made of stacked consecutive frames. The direct application of existing architectures on such inputs has shown mixed results. An alternative is to first extract optical flow or dense trajectories with an external algorithm \cite{simonyan2014,ye2015,wu2015} and feed the CNN with this information in addition to the pixel values. The success of this approach can be explained by the intrinsically different nature of spatial and temporal components, now separated during a preprocessing. The extraction of motion regardless of image contents is not trivial, and has been addressed by the long-standing line of successful optical flow algorithms. Conversely, the end-to-end training of CNNs on videos for high-level tasks has shown limited capability for identifying intermediate representations of motion. In this paper, we show that specifically training a CNN to extract optical flow can be achieved once some key principles are taken into account.

%The application of convolutional neural networks to videos has shown benefit from processing spatial and temporal information in separate streams. As an alternative to feeding the temporal stream with externally-computed optical flow
%this paper studies the extraction of motion within the network itself.
%They bear very different characteristics that call \eg for different forms of invariances that the CNN is expected to learn.
%The identification of motion independently from the spatial contents of the image involves a number invariances, in particular to phase, scale, and gradient orientation
%We derive an architecture capable of learning such invariances is derived from classical signal processing principles
%We train a CNN end-to-end to map a volume of pixels (a pair or series of frames) to its optical flow.
%We implement these operations as a CNN, and we learn spatiotemporal convolution kernels by standard backpropagation through this network. These convolution kernels decompose the video signal into frequency bands.

\begin{figure}[t]
  \begin{center}
  \includegraphics[width=0.93\linewidth,width=.99\linewidth]{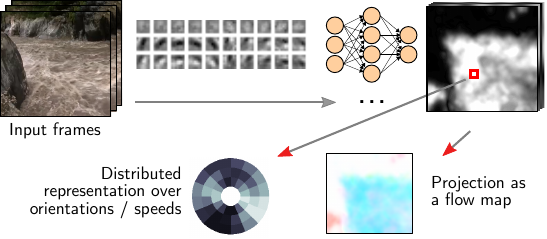}
  \end{center}
  \vspace{-6pt}
  \caption{The proposed CNN takes raw pixels as input and produces features representing evidence for motion at various speeds and orientations. These can be projected as a traditional optical flow map. First-layer kernels (pictured) typically identify translating patterns in the image.}
  %can describe multiple, overlapping motions at a single point
  \vspace{-9pt}
\end{figure}

We leverage signal processing principles and how motion manifests itself in the frequency domain of a spatiotemporal signal (\sect\ref{sect:principles}) to derive convolutions, pooling and non-linear operations able to map input pixels to a representation of motion (\sect\ref{ref:network}). The resulting network is designed as a building block for deeper architectures addressing higher-level tasks. The current practice of extracting optical flow as an independent preprocessing might be suboptimal, as the assumptions of an optical flow algorithm may not hold for a particular end application. The proposed approach would potentially allow to fine-tune the motion representation with the whole model in a deep learning setting.

The proposed network outputs a distributed representation of motion. The penultimate layer comprises, for each pixel, a population of neurons selective to various orientations and speeds. These can represent a multimodal distribution of activity at a single spatial location, and represent non-rigid, overlapping, and transparent motions that traditional optical flow usually cannot. The distributed representation can be used as a high-dimensional feature by subsequent applications, or decoded into a traditional map of the dominant flow. The latter allows training and evaluation with existing optical flow datasets.

The motivation for the proposed approach is not direct competition with existing optical flow algorithms. The aim is to enable using CNNs with videos in a more principled way, using pixel input for both spatial and temporal components. Instead of considering the flow of a complete scene as the end-goal, we rather wish to identify local motion cues without committing to an early scene interpretation. This contrasts with modern optical flow approaches, which often perform implicit or explicit tracking and/or segmentation. In particular, we avoid motion smoothness and rigidness priors, and spatial scene-level reasoning. This makes the features produced by the network suitable to characterize situations that break such assumptions, \eg with transparent phenomena and dynamic textures \cite{derpanis2012,teney2014}. As a downside, our evaluation on the Sintel benchmark shows inferior performance to state-of-the-art techniques. This confirms that a \emph{scene-level} interpretation of motion requires such priors and higher-level reasoning. In particular, we do not perform explicit feature tracking and long-range matching, which are the highlights of the best-performing methods on this dataset (\eg\cite{deepflow,brox2011}).
%on the order of classical optical flow methods

The contributions of this paper are fourfold. (1)~We derive, from signal processing principles, a CNN able to learn mapping pixels to optical flow. (2)~We train this CNN end-to-end on videos with ground truth flow, then demonstrate performance on the Middlebury benchmarks comparable to classical methods. (3)~We show how to enforce strict rotation invariance within a CNN through weight-sharing, and demonstrate significant benefit for training on a small dataset with no need for data augmentation. (4)~We show that the distributed representation of motion produced within our network is more versatile than traditional flow maps, able to represent phenomena such as dynamic textures and multiple, overlapping motions. Our pretrained network is available as a building block for deeper architectures addressing higher-level tasks. Its training is significantly less complex than competing methods \cite{flownet} while providing similar or superior accuracy on the Middlebury benchmark.

%- design network to extract optical flow with an architecture that can account for the required invariance to translation, phase and image texture
%- further enforce the desired invariances by regularizing the first-layer weights (spatiotemporal filters) via optimization with projected gradient descent to 
%- propose a novel form of in-network \emph{guided} pooling that pool feature maps while respecting natural image boundaries instead of a regular, fixed grid. This allows to recover, at the last layer of the network, a dense, high-resolution output
%We rather aim at providing a more general (\ie by handling multiple overlapping motions) low-level representation of motion, which can be used directly for various tasks, and that can also be optimized by fine-tuning for a particular application.

%====================================================================================================================
\section{Related work}

\shortTitle{Learning spatiotemporal features} Several recent works have used CNNs for classification and recognition in videos. Karpathy \etal~\cite{karpathy2014} consider the large-scale classification of videos, but obtain only a modest improvement over single-frame input. Simonyan and Zisserman \cite{simonyan2014} propose a CNN for activity recognition in which appearance and motion are processed in two separate streams. The temporal stream is fed with optical flow computed by a separate algorithm. The advantage of using separate processing of spatial and temporal information was further examined in \cite{wu2015,ye2015}. We propose to integrate the identification of motion into such networks, eliminating the need for a separate algorithm, and potentially allowing to fine-tune the representation of motion. Tran \etal~\cite{tran2014} proposed an architecture to extract general-purpose features from videos, which can be used for classification and recognition. Their deep network captures high-level concepts that integrate both motion and appearance. In comparison, our work focuses on the extraction of motion alone, \ie independently of appearance, as motivated by the two-stream approach \cite{simonyan2014}. Learning spatiotemporal features outside of CNN architectures was considered earlier. Le \etal \cite{le2011} used independent subspace analysis to identify filter-based features. Konda \etal~\cite{konda2013} learned motion filters together with depth from videos. Their model is based on the classical energy-based motion model, similarly to ours. Taylor \etal~\cite{taylor2010} used restricted Boltzmann machines to learn unsupervised motion features. Their representation can be used to derive the latent optical flow, but was shown to capture richer information, useful \eg for activity recognition. The high-dimensional features produced by our network bear similar benefits. Earlier work by Olshausen \etal~\cite{olshausen2003} learned sparse decompositions of video signals, identifying features resembling the filters learned in our approach. In comparison to all works mentioned above, we focus on the extraction of motion independently of spatial appearance, whereas decompositions such as in \cite{olshausen2003} result in representations that confound appearance and temporal information.

%Large-Scale Video Classification with Convolutional Neural Networks
%\cite{konda2013} Unsupervised learning of depth and motion

%However, for video classification, most deep network based approaches \cite{ji2010,karpathy2014,simonyan2014} demonstrated worse or similar results to the hand-engineered features [44]
%Some approaches \cite{ji2010,karpathy2014,simonyan2014} only focused on the static frames and short-term motion clues captured by a few adjacent frames
%most existing approaches fused the outputs of multiple networks in a very straightforward way \cite{simonyan2014}, which could lead to sub-optimal performance.
%Karparthy et al. compared several architectures for action recognition \cite{karpathy2014}.
%Tran et al. proposed to learn generic spatial-temporal features which can be computed efficiently \cite{tran2014}.
%\cite{simonyan2014} two-stream approach, two ConvNets trained to capture spatial and short-term motion, using frames / flows; final predictions obtained by averaging predictions of the 2 ConvNets

%-----------------------------------------------------------------
\shortTitle{Extraction of optical flow} The estimation of optical flow has been studied for several decades. The basis for many of today's methods dates back to the seminal work of Horn and Schunk \cite{hornSchunk}. The flow is computed as the minimizer of data and smoothness terms. The former relies on the conservation of a measureable image feature (typically corresponding to the assumption of brightness constancy) and the latter models priors such as motion smoothness. Many works proposed improvements to these two terms (see \cite{fortun2015} for a recent survey). Heeger \etal~\cite{heeger1987} proposed a completely different approach, applying spatiotemporal filters to the input frames to sample their frequency contents. This method naturally applies to sequences of more than two frames, and relies on a bank of hand-designed filters (typically, Gaussian derivatives or spatiotemporal Gabor filters \cite{adelson1985}). Subsequent improvements \cite{rust2006,solari2015,ulman2010} focused on the design of those filters. They must balance the sampling of narrow regions of the frequency spectrum, \ie, to accurately estimate motion speed and orientation, while retaining the ability to precisely locate the stimuli in image. A practical consequence of this tradeoff is the typically blurry flow maps produced the method. This historically played in favor of the more popular approach of Horn and Schunk. Another downside of the filter-based approach was the computational expense of convolutions. Our work revisits Heeger's approach, motivated by two key points. On the one hand, applying spatiotemporal filters naturally falls in the paradigm of convolutional neural networks, which are currently of particular for analyzing videos. On the other hand, modern advances can overcome the two initial burdens of the filter-based approach by (1) learning the filters using backpropagation, and (2) leveraging GPU implementations of convolutions.

Concurrently with our own developments, Fisher \etal~proposed another CNN-based method (Flownet \cite{flownet}). They obtain very good results on optical flow benchmarks. In comparison to our work, they train a much deeper network that requires tens of thousands of training images. Our architecture is derived from signal processing principles, which contains fewer weights by several orders of magnitude. This allows training on much smaller datasets. The final results in \cite{flownet} also include a variational refinement, essentially using the CNN to initialize a traditional flow estimation. Our procedure is entirely formulated as a CNN. This potentially allows fine-tuning when integrated into a deeper architecture.

A number of recent works \cite{derpanis2012,derpanis2010,teney2014,teney2015} studied the use of spatiotemporal filters to characterize motion in \eg transparent and semi-transparent phenomena, and dynamic textures such as a swirling cloud of smoke, reflections on water, or swaying vegetation. These works highlighted the potential of filter-based features, and the need for motion representations --~such as those produced by our approach~-- that go beyond displacement (flow) maps

\shortTitle{Invariances in CNNs} One of our technical contributions is to enforce rotation invariance by tiying groups of weights together. In contrast to weight \emph{sharing} which involves weights at different layers, this applies to weights of a same layer. Encouraging or enforcing invariances in neural networks has been approached in several ways. The convolutional paradigm ensure translation invariance by reusing weights between spatial locations. Other schemes of weight sharing were proposed in a simple model in \cite{fasel2006}, and later in \cite{le2010} with a method to learn which weights to share. No published work discussed the implementation of strict rotation invariance in CNN to our knowledge. In \cite{dieleman2015,laptev2015}, schemes akin to ensemble methods were proposed for invariance to geometric transformations. In \cite{rowley1998} and more recently in \cite{jaderberg2015}, a network first predicts a parametric transformation, used to rotate or warp the image to a canonical state before further processing. Our approach, in addition to the actual invariance, has the benefit of reducing the number of weights to learn and facilitates the training. Soft invariances, \eg to contrast can be encouraged by specific operations, such as local response normalization (LRN, \eg in \cite{krizhevsky2012}). We also use a number of such operations.
Note that this paper abuses of the term ``invariance'' for cases more accurately involving \emph{equi}variance or \emph{co}variance \cite{jayaraman2015}.

%\cite{kanazawa2014}
%\cite{oyallon2015}

%====================================================================================================================
\section{Filter-based motion estimation}
\label{sect:principles}

Our rationale for estimating motion with a convolutional architecture is based on the motion energy model \cite{adelson1985}. Classical implementations \cite{heeger1987,niyogi1995,ulman2010} are based on convolutions with hardcoded spatiotemporal filters (\eg 3D Gabors or Gaussian derivatives). The convolution of a signal with a kernel in the spatiotemporal domain corresponds to a multiplication of their spectra in the frequency domain. Convolutions with a bank of bandpass filters produce measurements of energy in these bands, which are then suitable for frequency analysis of the signal. A pattern moving in a video with a constant speed and orientation manifests itself as a plane in the frequency domain \cite{adelson1985,fleet1990}, and the signal energy entirely lies within this plane. It passes through the origin, and its tilt corresponds to the motion orientation and speed in the image domain. Classical implementations have used various schemes to identify the best-fitting plane to the energy measurements. In our model, we learn the spatiotemporal filters together with additional layers to decode their responses into the optical flow. Importantly, transparent patterns in an image moving with different directions or speeds correspond to distinct planes in the frequency domain. The same principle can thus identify multiple, overlapping motions.

%Why can handle multiple transparent motions while trained on single/rigid ones ? distributed representation at all stages, from motion evidence (motion filters), during processing all the way until before the output layer (regression/interpolation)

%In the case of videos, we use 3-dimensional kernels that are applied on volume of pixels of the video. Whereas previous work has studied the engineering of appropriate bandpass filters to decompose video signals, our work seeks to learn such filters. The motivation behind a learning approach is to allow optimizing and tuning the filters to particular domains.

%In comparison with classical implementation of the motion energy model, we learn convolution kernels from scratch with supervised training. 

%Importantly, each of these operations is motivated by desired invariances and well established signal processing principles. This contrasts with the common paradigm of deep architectures relying on large-scale training to learn those invariances. Practically, hard-coding relevant operations in the network does not require much stronger assumptions than artificially augmenting the training set as commonly done \cite{flownet}.

%====================================================================================================================
\section{Proposed network architecture}
\label{ref:network}

Our network is fully convolutional, \ie, it does not contain any fully connected layers. Each location in the output flow map is linked to a spatially-limited receptive field in the input, and each pixel of a training sequence can thus be seen as a unique training point. We describe below each layers of our network in their feedforward order. We use $x_{ijk}^{\lll}$ to refer to the scalar value at coordinates $(i,j,k)$ in the 3D tensor obtained by evaluating the $\lll$th layer. Indices $i$ and $j$ refer to spatial dimensions, $k$ to feature channels. We use a colon ($:$) to refer to all elements along a dimension. We denote with $\conv$ and $\convv$ convolutions in two and three dimensions, respectively. In contrast to CNNs used for image recognition, the desired output here is dense, \ie a 2D flow vector for every pixel. To achieve this, all convolutions use a 1\px stride and the pooling (\eq\ref{eq:pooling}) a 2\px stride, all with appropriate padding. The output is thus at half the resolution of the input. Our experiments use bilinear upsampling (except otherwise noted) to obtain flow fields at the original resolution.

%(explaining in part the limited need for training data)

%-----------------------------------------------------------------
\subsection{Network input}
The input to the network is the $H\n\times\n W \n\times\n F$ volume of pixels made by stacking $F$ successive grayscale frames of a video (typically, $F\mm=\mm3$). The desired network output is the flow between frames $\lceil F/2 \rceil$ and $\lceil F/2 \rceil\n+\n1$.
\abovedisplayskip=2pt
\belowdisplayskip=2pt
\begin{gather} \label{eq:input}
  x_{::k}^{1} ~=~ k \textrm{th frame of the video} ~~~~\forall\; k\m=\m1...F.
\end{gather}

%-----------------------------------------------------------------
%\subsection{Invariance to additive brightness and local contrast}
\subsection{Invariance to brightness and contrast}

The estimated motion should be insensitive to additive changes of brightness of the input. Since the subsequent processing will be local, instead of subtracting the average brightness over the whole image, we subtract the local low-frequency component:
\begin{gather} \label{eq:centerSurround}
  x_{::k}^{\ll} ~=~ x_{::k}^{\lll} ~-~ (x_{::k}^{\lll} ~\conv~ \mathcal{H}^\gaussKernels)  ~~~~\forall\, k,
\end{gather}
where $\mathcal{H}^\gaussKernels$ denotes a fixed, 2D Gaussian kernel of standard deviation $w/3$. Note that this operation could also be written as a convolution with a center-surround filter. We then ensure invariance to local constrast changes with a normalization using the standard deviation in local neighbourhoods:
\begin{gather} \label{eq:stdNormalization}
  x_{ijk}^{\ll} ~=~ x_{ijk}^{\lll} ~~/~~ \stddev_{\vspace{8pt}\m i',j' \in \Omega(i,j)}\big( x_{i'j'k}^{\lll}\big ) ~~~~\forall\, i,j,k,
\end{gather}
where $\Omega(i,j)$ refers to the square region of length~$w$ centered on $(i,j)$. 

The two above operations proved essential to learn subsequent filters with little training data. They can be seen as a local equivalent to a typical image-wide whitening \cite{lecun1998b}. This formulation is better suited to the subsequent local processing of a fully convolutional network.

%-----------------------------------------------------------------
\subsection{Motion detection}
This key operation convolves the volume of pixels with learned 3D kernels. They can be interpreted as spatiotemporal filters that respond to various patterns moving at different speeds.
\begin{gather} \label{eq:motionFilters}
  x_{::k}^{\ll} ~=~ x_{:::}^{\lll} ~\convv~ \mathcal{H}_k^\motionKernels + {b}_k^\motionKernels ~~~\forall~k\m=\m1...MO~~,
\end{gather}
where $\mathcal{H}_k^\motionKernels$ are $MO$ learned convolution kernels of size $w \m\times\m w \m\times\m F$, and ${b}_k^\motionKernels$ the associated biases. The constants $M$ and $O$ respecitvely fix the number of independent kernels and the number of orientations explicitly represented within the network (\sect\ref{sect:orientationInvariance}).
%Typically, $w\m=\m7, M\m=\m4, O\m=\m12$.

\begin{figure}[t]
  \begin{center}
  \includegraphics[width=0.92\linewidth]{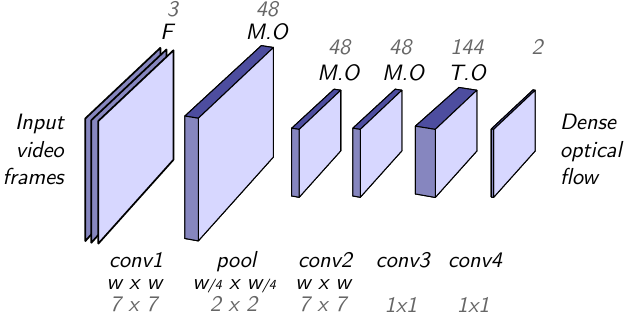}
  \end{center}
  \vspace{-4pt}
  \caption{The proposed neural network takes raw pixels of $F$ video frames as input, and outputs a dense optical flow at half its resolution. The network comprises 2 convolutional layers, 1 pooling layer, and 2 pixelwise weights ($1\n\times\n 1$ convolutions). Dimensions of receptive fields and feature maps are shown, with numbers chosen in our implementation in gray. Normalizations and non-linearities are not shown.}
  \label{fig:architecture}
  \vspace{-9pt}
\end{figure}

%-----------------------------------------------------------------
\subsection{Invariance to local image phase}
The learned kernels used as motion detectors above typically respond to lines and edges in the image. The estimated motion should however be independent of such image structure. Classical models \cite{heeger1987,solari2015} account for this using pairs of quadrature filters, though this is not trivial to enforce with our learned filters. Instead, we approximate a phase-invariant reponse as follows. (1) The response of the convolution is rectified by pointwise squaring. Responses out of phase by 180\degree (\eg dark-bright and bright-dark transitions) then give a same response. (2) We apply a spatial max-pooling. The wavenumber of the pattern captured by our kernels of size $w$ is at least $2/w$ cycles/px. The worst-case phase shift of 90\degree then corresponds to a spatial shift of $w/4$ px. We maxpool responses over windows of size $w/4$ with a fixed stride of $2$, and thus approximate a phase-invariant response at the price of a lowered resolution.
\begin{gather} \label{eq:pooling}
  x_{ijk}^{\ll} ~=~ \max_{\vspace{8pt}\m i',j' \in \Omega'(2i,2j)}\big( {x_{i'j'k}^{\lll}}^2 \big) ~~~~\forall\, i,j,k,
\end{gather}
where $\Omega'(2i,2j)$ refers to the square region of length~$\lceil w/4 \rceil$ centered on $(2i,2j)$.

%-----------------------------------------------------------------
\subsection{Invariance to local image structure}
The estimated motion should be independent from the amount and type of texture in the image. To account for intensity differences of patterns at different orientations at any particular location (\eg a grid pattern of horizontal lines crossing fainter vertical ones), we normalize the responses by their sum over all orientations \cite{heeger1987}:
\begin{gather} \label{eq:normalization}
  x_{ijk}^{\ll} ~=~ x_{ijk}^{\lll} \Big/ \big(\textstyle \sum_{k'} {x_{ijk'}^{\lll}} + \epsilon \big)~~~\forall\, i,j,k,
\end{gather}
where $\epsilon$ is a small constant to avoid divisions by a small value in low-texture areas of the image. The sum is performed over feature channels $k'$ that correspond to the $O$ variants of $k$ at all orientations (see \sect\ref{sect:orientationInvariance})

%for non-uniform presence of patterns at all orientations, thereby accounting for the well-known aperture problem, 
To account for the aperture problem, we allow local interacton by introducing an additional convolutional layer with $MO$ learned kernels $\mathcal{H}^\poolingKernels$ of size $w \m\times\m w \times MO$.
\begin{gather} \label{eq:smoothing}
  x_{::k}^{\ll} ~=~ x_{:::}^{\lll} ~\convv~ \mathcal{H}_k^\poolingKernels + {b}_k^\poolingKernels ~~~\forall~k\m=\m1...MO~~,\\
  x_{ijk}^{\ll} ~=~ \max(x_{ijk}^{\lll}, 0)~~.
\end{gather}
The classical hardcoded models typically use here 2D convolutions with Gaussian kernels. Our experiments showed that supervised training lead to similar kernels, although slightly non-isotropic, and modeling some cross-channel and center-surround interactions.
%Such interactions have been discussed in studies of motion perception in area MT \cite{??}.

%-----------------------------------------------------------------
\subsection{Decoding into flow vectors}
The features maps at this point represent evidence for different types of motion at every pixel. This evidence is now decoded with a hidden layer, a softmax nonlinearity and a linear output layer:
\begin{gather}
  \label{eq:decoding} x_{::k}^{\ll} ~=~ x_{:::}^{\lll} ~\convv~ \mathcal{H}_k^\decodingKernels + {b}_k^\decodingKernels ~~~\forall~k\m=\m1...TO\\
  \label{eq:softmax} x_{ijk}^{\ll} ~=~ e^{x_{ijk}^{\lll}} \big/ \textstyle \sum_{k'} e^{x_{ijk'}^{\lll}} ~~~~\forall~i,j,k\m=\m1...TO\\
  \label{eq:output} x_{::k}^{\ll} ~=~ x_{:::}^{\lll} ~\convv~ \mathcal{H}_k^\outputKernels + {b}_k^\outputKernels ~~~\forall~k\m=\m\{1,2\},
\end{gather}
%We learn $NO$ weights $mathcal{H}_k^\decodingKernels$ and biases $b_k^\decodingKernels$ and $\mathcal{H}_k^\outputKernels$ are $1\n\times\n1$.
where $T$ is a constant that fixes the number of hidden units. The decoding is performed pixelwise, \ie $\mathcal{H}_k^\decodingKernels$ and $\mathcal{H}_k^\decodingKernels$ are $1\m\times\m 1$.
Intuitively, the activations of the hidden layer (\eq\ref{eq:decoding}) represent scores for motions at $S$ and $O$ discrete speeds and orientations, of which the softmax picks out the highest. Assuming a unimodal distribution of 
scores (\ie a single motion at any pixel), the output layer interpolates these scores and maps them to a 2D flow vector for every pixel (see \sect\ref{sect:training}).

%-----------------------------------------------------------------
\subsection{Invariance to in-plane rotations}
\label{sect:orientationInvariance}
In our context, rotation invariance implies that a rotated input must produce a correspondingly rotated output. Note the contrast with image recognition where rotated inputs should give a \emph{same} output. All of our learned weights (\eq\ref{eq:motionFilters}--\ref{eq:decoding}) are split into groups corresponding to discrete orientations. The key is to enforce these groups of weights to be equivalent, \ie so that they make the same use of features from the preceding layer at the same \emph{relative} orientations. In addition, convolutional kernels need to be 2D rotations of each other. These strict requirements allow us to maintain only a single version of the weights at a canonical orientation, and generate the others when evaluating the network (see \fig\ref{fig:weights}). During training with backpropagation, the gradients are aligned with this canonical orientation and averaged to update the single version of the weights.

Formally, let us consider a convolutional layer\footnote{The general formulation applies to convolutional layers as well as to our pixelwise weights (\eq\ref{eq:decoding}, \ref{eq:output}), in which case the 2D rotation of the kernel has no effect.} $\ll$. The feature maps $x^{\lll}$ (respectively $x^{\ll}$) are split into $O$ ($P$) groups of $M$ ($N$) channels. For example, in \eq\ref{eq:smoothing}, $O\n=\n P$ and $M\n=\n N$. The groups of channels correspond to regular orientations $\thetaIn_{i}$ ($\thetaOut_{j}$) in $[0,2\pi[$. Considering the convolution weights $\mathcal{H}$ and their slice $h_{imjn}$ the 2D kernel acting on the input (respectively output) channel of orientation $\thetaIn_{i}$ ($\thetaOut_{j}$), we constrain the weights as follows:
\begin{gather}
  \label{eq:rotationTies1}
  h_{imjn} ~=~ \rot _{\thetaOut_{j}-\thetaOut_{j'}} \big( h_{i'mj'n} \big) 
  \\
  \label{eq:rotationTies2}
  \forall ~i,i',j,j',m,n~~\textrm{s.t.}~~cos(\thetaOut_{j'}-\thetaIn_{i'})=cos(\thetaOut_{j}-\thetaIn_{i})
\end{gather}
\eq\ref{eq:rotationTies1} ensures that convolution kernels are rotated versions of each other (implemented with bilinear interpolation) and \eq\ref{eq:rotationTies2} ensures that the same weights are applied to input channels representing a same \emph{relative} orientation with respect to a given output channel of the layer. In other words, the weights are shifted between each group so as to act similarly on channels representing the same relative orientations (see \fig\ref{fig:weights}). In \eq\ref{eq:smoothing} and \ref{eq:decoding}, $N\m=\m N'$. In \eq\ref{eq:motionFilters}, $N\m=\m1$. In \eq\ref{eq:output}, $N\m=\m2$ with $\thetaOut=\{0,\pi/2\}$.

%It has with  input and $SN'$ output channels (in this example $N\n=\n N'$). 
%(\eq\ref{eq:motionFilters}, \ref{eq:smoothing}, \ref{eq:decoding} or \ref{eq:output})
%The approach can be seen as a strong regularizer on the weights of the network, enforcing a motion in a given orientation to be decoded the same way as motion in another orientation.

It follows that the number of convolution kernels to explicitly maintain is reduced from $OPMN$ to $\lceil O/2 \rceil MN$. In \eq\ref{eq:smoothing} for example, in our implementation with $O\n=\n P\n=\n 12$ orientations, this amounts to a decrease by a factor 24. It allows training on small amounts of data with lower risk of overfitting. This also negates the need to artificially augment the dataset with rotations and flips.

%Other way to explain orientation sharing: use theta=relative orientation of channels (2*pi/O); then separate 2 enforcements: rotate2D / shift weights

%====================================================================================================================
\section{Multiscale processing}

\begin{figure}[t]
  \begin{center}
  \includegraphics[width=0.95\linewidth,width=.95\linewidth]{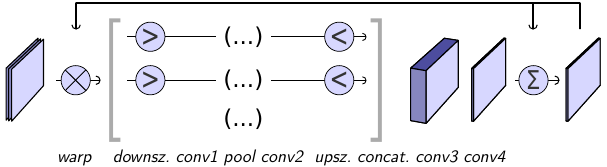}
  \end{center}
  \vspace{-5pt}
  \caption{We bring two modifications to the basic network (\fig\ref{fig:architecture}) for multiscale processing. First, we apply the network on several downsampled versions of the input, concatenating the feature maps before the decoding stage. Second, we add a recurrent connection to warp the input frames according to the estimated flow, iterating its evaluation for a fixed number of steps. This is inspired by the classical ``coarse-to-fine'' strategy for optical flow, and designed to estimate motions larger than the receptive field of the network output units.}
  \label{fig:multiscale}
  \vspace{-9pt}
\end{figure}

Equations \ref{eq:input}--\ref{eq:output} form a complete network that maps pixels to a dense flow field. However, the detection of large motion speeds is limited by the small effective receptive field of the output units, due to the limited number of convolutional layers. We remedy this in two ways (\fig\ref{fig:multiscale}) without increasing the number of weights to train. First, we apply the network (\eq\ref{eq:input}--\ref{eq:smoothing}) on multiple downsized versions of the input frames. The feature maps are brought back to a common resolution by bilinear upsampling and concatenated before the decoding stage (\eq\ref{eq:decoding}--\ref{eq:output}).
%, using the invariance to scale of the motion estimation principles (\sect\ref{sect:principles})

Second, we add a recurrent connection to the network that warps the input frames according to the current estimate of the flow. The evaluation runs through the recurrent connection for a fixed number of steps. This is inspired by the classical coarse-to-fine warping strategy \cite{memin1998}. It allows the model to approximate the flow iteratively. Note that the recurrent connection to the warping layer is not a strictly linear operation contrary to typical recurrent neural networks. Training is performed by backpropagation through the unfolded recurrent iterations, but not through the recurrent connection itself. Since the result of each iteration is summed to give the final output, we can append and sum the same loss (\sect\ref{sect:training}) at the end of each unfolded iteration.
%A comparable principle for estimating a transformation applied back to the input was recently discussed in \cite{jaderberg2015}.

%normalization over orientations; generates a signal that is independent of local brightness changes, which has been previously used in motion detectors (Heeger, 1987; Bayerl and Neumann, 2004)."
% 2012 Modeling Binocular and Motion Transparency Processing by Local Center-Surround Interactions
% "motion signals are represented in populations of neurons, selective to disparity and motion, respectively. These populations can represent multiple activations at a single spatial location through the formation of multiple local peaks in the activity distribution"
%cite: Filter selection model for motion segmentation and velocity integration, Steven J. Nowlan* and Terrence J. Sejnowski
%maybe add some of their justifications to our technical section (with citation !)

%====================================================================================================================
\section{Implementation and training}
\label{sect:training}

%\subsection{Loss function}
We train the weights ($\mathcal{H}^\motionKernels, ..., \mathcal{H}^\outputKernels$) and biases (${b}^\motionKernels, ..., {b}^\outputKernels$) end-to-end with backpropagation. Even though the network ultimately performs a regression to flow vectors, we found more effective to train it first for \emph{classification}. First, we pick a number $TO$ of flow vectors uniformly in the distribution of training flows. We train the network for nearest-neighbour classification over these possible outputs, with a logarithmic loss over \eq\ref{eq:softmax}. Second, we add the linear output layer (\eq\ref{eq:output}) and initialize the rows of $\mathcal{H}_k^\outputKernels$ with the vectors used for classification. Second, the network is fine-tuned with a Euclidean (``end-point error'') loss over the decoded vectors. That fine-tuning has a marginal effect in practice. The softmax values are practically unimodal and sum to one by construction, hence they can directly interpolate the vectors used for classification with a linear operation. We believe the 2-step training is helpful because the Euclidean loss alone does not reflect well the quality (\eg smoothness) of the flow. The classification loss guides the optimization towards a better optimum.
%and zero biases

Considering the above, the softmax values (\eq\ref{eq:softmax}) constitute a distributed representation of motion, where each dimension corresponds to a different orientation and speed. Feature maps at this layer can encode multimodal distributions over this representation and represent patterns that optical flow cannot.

\begin{figure}[t]
  \begin{center}
  \scriptsize
  %\vspace{-30pt}
  \renewcommand{\arraystretch}{0.8}
  \renewcommand{\tabcolsep}{0.60mm}
  \begin{tabular}{ccc}
    \hangBox{{\includegraphics[height=17.5mm]{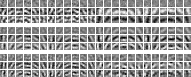}}} &
    \hangBox{{\includegraphics[height=17.5mm]{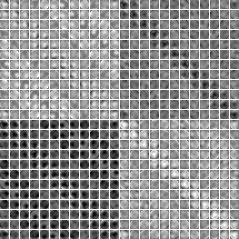}}} &
    \hangBox{{\includegraphics[height=17.5mm]{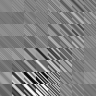}}}
  \end{tabular}
  \end{center}
  \vspace{-7pt}
  \caption{Convolutional kernels $\mathcal{H}^\motionKernels$, $\mathcal{H}^\poolingKernels$, and pixelwise weights $\mathcal{H}^\decodingKernels$ learned on the Middlebury dataset (selection). The obvious structure is enforced by constraints that ensure rotation invariance of the overall network.}
  \label{fig:weights}
  \vspace{-12pt}
\end{figure}

%====================================================================================================================
\section{Experimental evaluation}
\label{sect:experiments}

We present three sets of experiments. (1)~We evaluate the non-standard design choices of the proposed architecture through ablative analysis. (2)~We compare our performance versus existing methods on the Middlebury and Sintel benchmarks. (3)~We demonstrate applicability to dynamic textures and multiple transparent motions that goes beyond traditional optical flow. Code and trained models are available on the author's website \cite{myWebsite0}.

%====================================================================================================================
\subsection{Ablative analysis}
\label{sec:ablative}

Our ablative analysis uses the public Middlebury dataset, with the sequences split into halves for training (Grove2, RubberWhale, Urban3) and testing (Grove3, Dimetrodon, Hydrangea). For each run, we modify the network in one particular way, retrain it from scratch, and report its performance on the test set (Table~\ref{tab:ablative}). We observe that all preprocessing and normalization steps (2--3) have a positive impact, and some are even necessary for the training to converge at all (at least with the small dataset used in these experiments). Operations inspired by the classical implementations of the motion energy model, \ie, (8)~rectification by squaring of filter responses and (6)~normalization across orientations, proved beneficial as well. This shows the benefit for our principled approach to network design. We perform a comparison to the classical, hardcoded filter-approach approach \cite{heeger1987} by setting the first filters to Gaussian derivatives (4). Although learned filters are visually somewhat similar (\fig\ref{fig:weights}), learned kernels clearly perform better. This confirms the general benefit of a data-driven approach to motion estimation. 

%Please consult Table~\ref{tab:ablative} for additional comparisons.
%Graph of EPE on validation set during training
%	for random output layer + EPE
%	for fixed output layer + EPE
%	for fixed output layer + softmax logloss THEN EPE (better)
%sharing weights between all versions of the network applied at the different scales: effect proved mitigated due to, we believe, dataset-specific considerations.

\begin{table}
  \scriptsize
  \renewcommand{\tabcolsep}{0.1mm}
  \renewcommand{\arraystretch}{1.0}
  \begin{center}
  %\resizebox{\linewidth}{!}{ \begin{tabular}{lcc}
  %\begin{tabular*}{\textwidth}{@{}>{\bfseries}l@{\extracolsep{\fill}}*{9}{c}@{}}
  \begin{tabular*} {\linewidth} {@{}>{}l@{\extracolsep{\fill}}*{2}{c}@{}}
  %~ & EPE & AAE \\ ~ & (\px) & (deg) \\
  ~ & EPE (\m\px\m) & \gt{AAE (\degreeSymbol)} \\
  \toprule\vspace{-10pt}\\
  %\hline
  Full model, $F\mm=\mm3$, $O\mm=\mm12$, 10 scales, 1 rec.iter. & 0.66 & \gt{6.9} \\
  \hline
  \tiny{(1)}\footnotesize~Number of input frames $F\mm=\mm2/5$ & 0.67~0.90 & \gt{6.8~8.7} \\
  \hline
  \tiny{(2)}\footnotesize~No center-surround filter & \nc & \gt{\nc} \\
  \tiny{(3)}\footnotesize~No local normalization & 0.67 & \gt{6.9} \\
  \hline
  \tiny{(4)}\footnotesize~Hard-coded $\mathcal{H}_k^\motionKernels$: Gauss. deriv. & 0.78 & \gt{8.4} \\
  \hline
  \tiny{(5)}\footnotesize~No L1-normalization over orientations & \nc & \gt{\nc} \\
  \tiny{(6)}\footnotesize~No pooling for phase invariance & 0.93 & \gt{11.2} \\
  \tiny{(7)}\footnotesize~ReLU after conv1 (default: squaring) & \nc & \gt{\nc} \\
  \hline
  \tiny{(8)}\footnotesize~No constraints for rotation invariance & \nc & \gt{\nc} \\
  \tiny{(9)}\footnotesize~Number of orientations $O\mm=\mm6/8/16$ & 1.65~0.72~0.65 & \gt{26.6~8.1~6.6} \\
  \hline
  \tiny{(10)}\footnotesize~Loss: classification/regression  & 0.66/0.83 & \gt{7.0~8.7} \\
  \multicolumn{3}{l}{~~~~\,(default: 2-step, classification (logarithmic) then regression (Euclidean))}\\
  \hline
  \tiny{(11)}\footnotesize~Number of scales: 4/8/16 & 0.69~0.66~0.67 & \gt{6.9~6.8~6.9} \\
  \hline
  \tiny{(12)}\footnotesize~Recurrent iterations: 2/3/4/5 & 0.57~0.55~0.56~0.57 & \gt{6.5~6.4~6.7~6.8} \\
  \toprule\vspace{-11pt}\\
  %\hline
  \end{tabular*}
  %\end{tabular} }
  \end{center}
  \vspace{-6pt}
  \caption{We evaluate every non-standard design choice by retraining a modified network. \nc~denotes networks for which the training did not converge within a reasonable number of iterations. See discussion in \sect\ref{sec:ablative}.}
  \label{tab:ablative}
  \vspace{-7pt}
\end{table}

\begin{table}
  \scriptsize
  %\tiny
  \renewcommand{\tabcolsep}{0.08mm}
  \renewcommand{\arraystretch}{1.0}
  \begin{center}
  %\resizebox{\linewidth}{!}{ \begin{tabular}{l|cc|cc|c}
  %\begin{tabular*} {\textwidth} {@{}>{}l@{\extracolsep{\fill}}*{6}{c}@{}}
  \begin{tabular*} {\linewidth} {@{}>{}l@{\extracolsep{\fill}}*{8}{c}@{}}
  ~      & \multicolumn{2}{c}{Middlebury}	& \multicolumn{2}{c}{Sintel private} & Time \\
  Method & public		& private		& clean & final	& (sec) \\
  \toprule\vspace{-11pt}\\
  %\hline
  FlowNetS \cite{flownet}             	& 1.09	\gt{13.28} & --		& 4.44 & 7.76 & 0.08 \\
  FlowNetS + refinement \cite{flownet}		& 0.33	\gt{3.87}& 0.47	\gt{4.58}& 4.07 & 7.22 & 1.05 \\
  DeepFlow \cite{deepflow}				& 0.21	\gt{3.24}& 0.42	\gt{4.22}& 4.56 & 7.21 & 17 \\
  LDOF \cite{ldofflow}       			& 0.45	\gt{4.97}& 0.56	\gt{4.55}& 6.42 & 9.12 & 2.5 \\
  Classic++ \cite{sun2010}   			& 0.28	\gt{~~~--~~~}		& 0.41	\gt{3.92}	& 8.72 & 9.96 & -- \\
  FFV1MT \cite{solari2015}				& 0.95	\gt{9.96}	& 1.24	\gt{11.66}& -- & -- & -- \\
  \toprule\vspace{-11pt}\\
  Proposed                            	& (0.45)	\gt{(5.47)}	& 0.70	\gt{6.41}& 9.36 & 10.04 & 6 \\
  Proposed + refinement as \cite{flownet}     	& (0.35)	\gt{(4.10)}	& 0.58	\gt{5.22}& 9.47 & 10.14 & 7 \\ 
  \hline
  %\bottomrule
  \toprule\vspace{-11pt}\\
  \end{tabular*}
  %\end{tabular} }
  \end{center}
  \vspace{-6pt}
  \caption{Comparison with existing algorithms on the Middlebury and Sintel benchmarks. We report average end-point errors (EPE, in pixels) average angular errors (AAE, in gray, in degrees), and execution times per frame on Sintel. Numbers in parentheses correspond to the sets used for training the model.}
  \label{tab:sintel}
  \vspace{-12pt}
\end{table}

%\degreeSymbol
%\m\px\m

%====================================================================================================================
\subsection{Performance on optical flow benchmarks}

We use the Middlebury and Sintel benchmarks for evaluation of networks trained from scratch on their respective training sets. A network trained on the smaller Middlebury dataset performs decently on Sintel sequences with small motions. However, most include much faster motions that had to be retrained for.

\begin{figure}[h]
  \vspace{-3pt}
  \scriptsize
  \renewcommand{\tabcolsep}{0.3mm}
  \renewcommand{\arraystretch}{0.8}
  \begin{center}
  %\resizebox{\linewidth}{!}{ \begin{tabular}{cccc}
  \begin{tabular}{cccc}
  \vspace{1pt}Input & Ground truth & As proposed & Flownet \cite{flownet} \\
  \middleburyRow{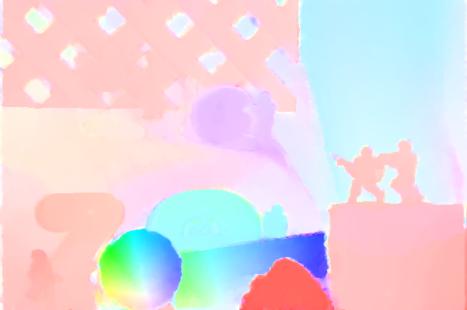}
  \middleburyRow{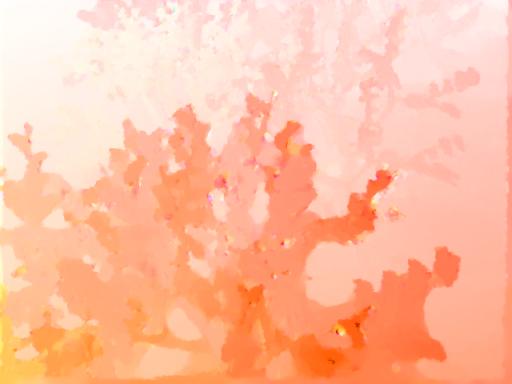}
  \middleburyRow{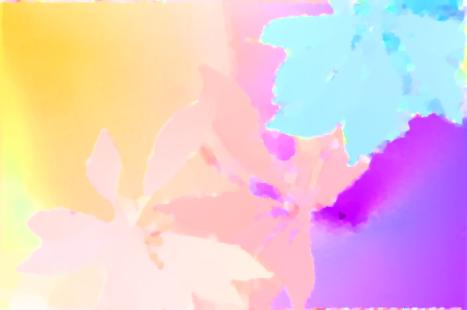}
  \middleburyRow{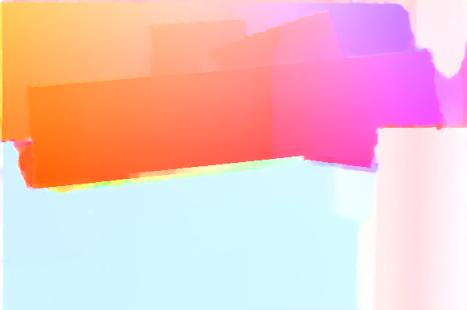}
  \middleburyRow{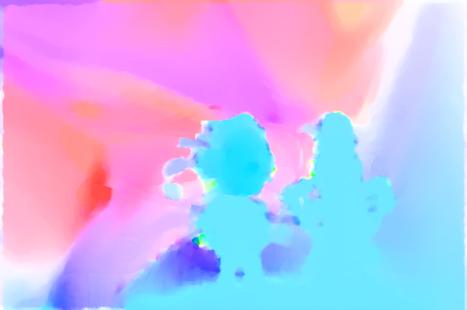}
  %\middleburyRow{6}
  %\middleburyRow{7}
  %\middleburyRow{8}
  \sintelRow{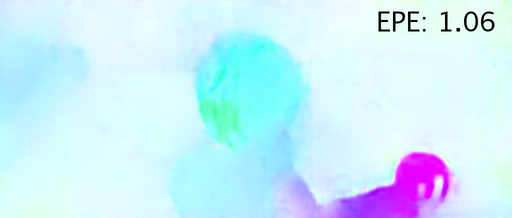}
  %\sintelRow{2}
  %\sintelRow{3}
  %\sintelRow{4}
  \sintelRow{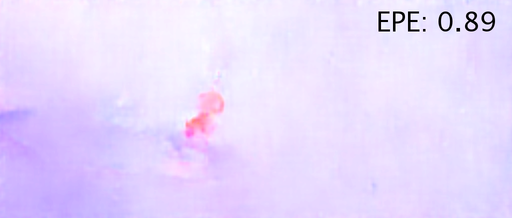}
  \vspace{3pt}\sintelRow{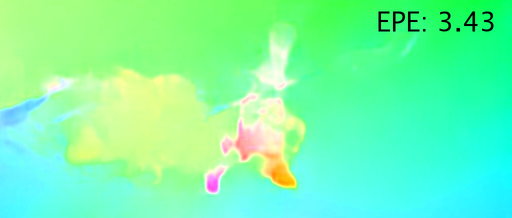}
  %\sintelRow{7}
  %\sintelRow{8}
  %\sintelRow{9}
  %\sintelRow{10}
  %\sintelRow{11}
  \vspace{1pt}Input & Ground truth & As proposed & Occlusions \\
  \sintelRowSimple{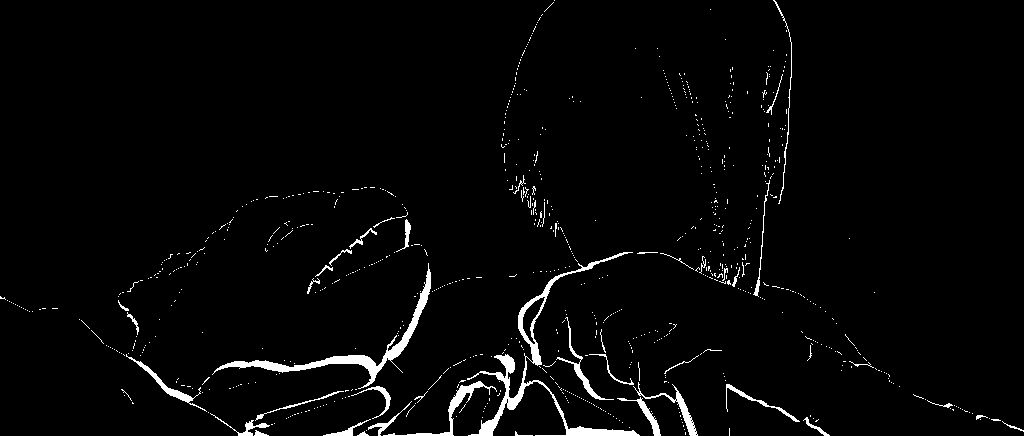}
  \sintelRowSimple{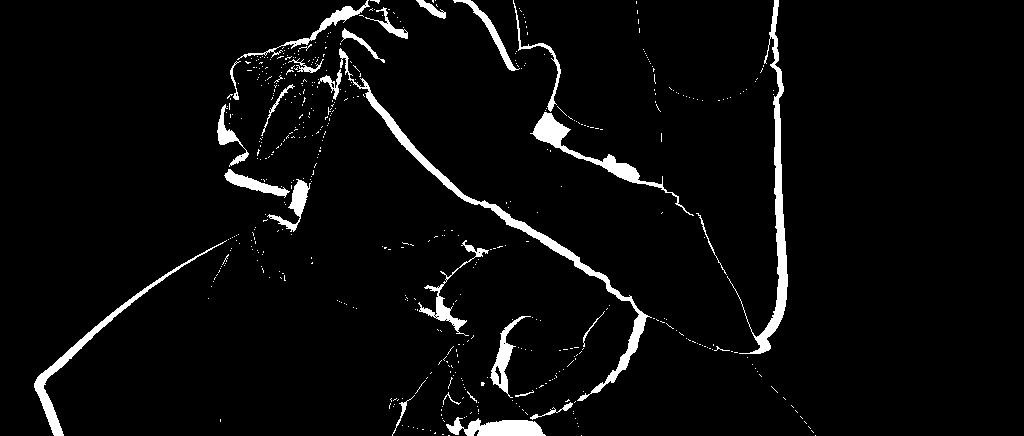}
  \sintelRowSimple{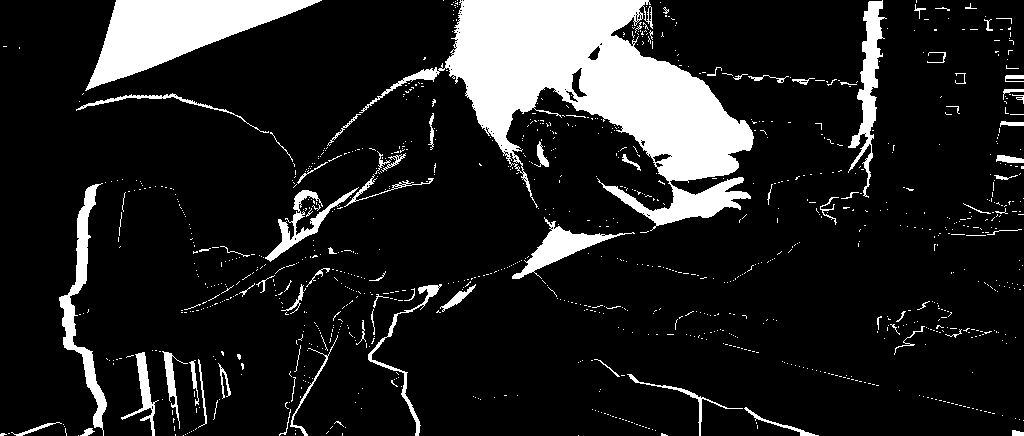}
  \sintelRowSimple{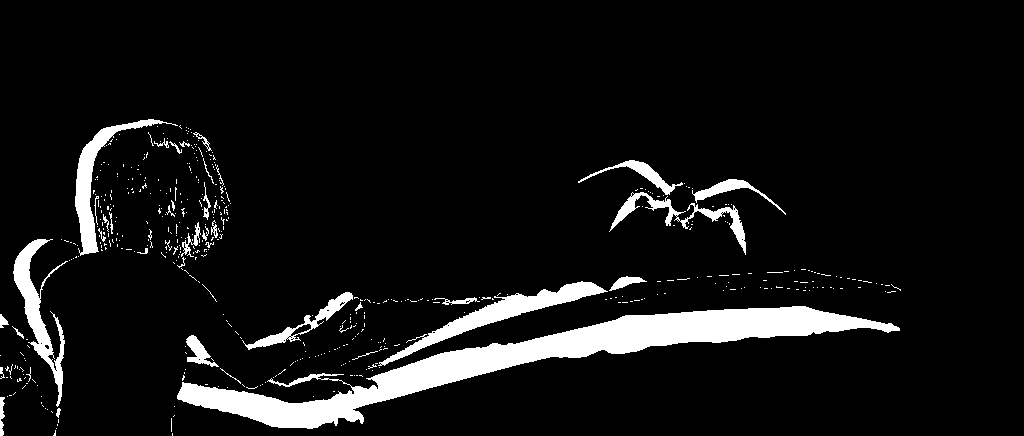}
  \end{tabular}
  \end{center}
  \vspace{-6pt}
  \caption{Estimated flow on sequences from the Middlebury (top five) and Sintel (others) datasets. Most failure cases (\eg bottom two) occur near (dis)occlusions, which are common on the Sintel dataset due to large motions of small objects and the small relative camera field-of-view.}
  \label{fig:sintel}
  \label{fig:middlebury}
  \vspace{-5pt}
\end{figure}

\shortTitle{Middlebury} Flow maps estimated by our method on the Middlebury dataset \cite{middlebury} are generally smooth and accurate (see \fig\ref{fig:middlebury}). Most errors occur near boundaries of
objects that become, or cause occlusions. Altough our flow maps remain generally more blurry than those of state-of-the-art methods, some fine details are remarquably well preserved (\eg \fig\ref{fig:middlebury}, second row). This blurriness, or imprecision in the spatial localization of motions, is a well-known drawback of filter-based motion estimation. Convolutional kernels of smaller extent would be desirable to provide better localization, but the extent of its response in the frequency domain (\sect\ref{sect:principles}) would correspondingly increase, which would imply a coarser sampling of the signal spectrum and lesser accuracy in motion direction and speed. Comparisons with existing methods show performance on the level of classical methods. We obtain much better performance than the recent implementation of Solari \etal~\cite{solari2015} of a filter-based method with no learning.

%In comparison to the CNN-based method Flownet \cite{flownet}, we obtain much better results out of the CNN. They also propose a varational refinement step, essentially using the CNN to initialize a traditional flow estimation. For a complete comparison, we applied a similar post-processing to our results, and then obtained very similar results to theirs.
%and strong gradients, where the relatively large spatial extent of our first-layer filters ($7\m\times\m 7$\px) produces features that can ``spill'' on either side of the gradient.
%Note that more recent, well-engineered algorithms have given more accurate results on this benchmark, and perform thus better than the proposed method on this particular type of scenes. Interestingly however, we perform much better than the best reported filter-based implementation to our knowledge \cite{solari2015}. An ablative analysis of or contributions and of various parameters of our method is provided in Table~\label{tab:middlebury}.
%Some inaccuracies occur in areas of low texture. These are well-known challenges for optical flow algorithms, that are typically solved by piecewise-smoothness priors. Our model does not include such explicit priors. 
%Discussion: perf on tr/te (limited by capactity of model, not learning or amount of data, < our choice of extracting local motion cues but not do scene-level reasoning), small improvement with refinement
%For full comparison with \cite{flownet}, we also processed our results with their variational refinement (without the use of image boundaries, as they reported it not to have a large effect).

\shortTitle{Sintel} The MPI Sintel dataset \cite{sintel} contains computer-generated scenes of a movie provided in ``clean'' and ``final'' versions, the latter including atmospheric effects, reflections, and defocus/motion blur. Flow maps estimated by our method (\fig\ref{fig:sintel}) are often smooth. Flows in scenes with small motions are usually accurate, but they lack details at the objects' borders and near small image details. Although this is partly alleviated by our recurrent iterative processing within the network, large errors remain in scenes with fast motions. This is reflected by a poor quantitative performance (Table~\ref{tab:sintel}). Additional insights can be gained by examining the flow maps (\fig\ref{fig:sintel}). Errors arise not on the estimates of large motions, but on their localization, in particular near zones of (dis)occlusions. This is obvious \eg in \fig\ref{fig:sintel}, last row, with a thin wing flapping over a blank sky. Although the actual motion (in yellow) is detected, it spills on both sides of the thin wing structure. Since such occlusons are caused by large motions, they result in a large penalty in EPE. As argued before \cite{sintel}, good overall performance in such situations clearly require reasoning over larger spatial and/or temporal extent than the local motions cues that our method was designed around.

Interesting comparisons can be made with the competing approach Flownet \cite{flownet}. It uses a more standard deep architecture with numerous convolutional and pooling layers. It also includes a variational refinement to improve the precision of motion estimates from the CNN. As discussed in \cite{flownet}, this refinement cannot correct large errors of correspondingly large motions. For comparison, we applied this same post-processing to our own results. Our results right off the CNN on Middlebury are already accurate, and the post-processing brings only marginal improvement (Table~\ref{tab:sintel}). The refinement on Flownet has a stronger effect: the output of their CNN is less precise, and it benefits more from this refinement. Looking at the Sintel dataset, the situation is very different. The main metric (the average EPE) is dominated by large motions, which are the weak point of the filter-based principles (\sect\ref{sect:principles}) that we rely on. Flownet is particularly good at long-range matching thanks to its deep architecture, and this results in vastly superior performance. As stated above, the refinement is of little use with large motions, and brings minimal improvement to either method on Sintel. In conclusion, the different design choices in Flownet and our approach seem complimentary in different regimes. It would be interesting to investigate how to combine their strengths.

%occlusion areas = where the optical flow model of image displacements is not sufficient
%This is an inherent limitation of the multiscale approach that we adopted: the coarser scales, used to identify faster motion, do not include fine image details, which can then be missed from the estimated motion. 
%The quantitative error (\tab\ref{tab:sintel}) is higher than methods specifically designed to handle such issues.
%Both on Middlebury and Sintel, the vast majority of our errors occur at (dis)occlusions. This is an inherent limitation of the motion estimation principles discussed in \sect\ref{sect:principles}. Occlusions of large areas are very common on Sintel, due to large motions of thin structures. The relative camera field-of-view is small, and large parts of objects commonly disappear from one frame to the next. Some optical flow algorithms can handle this \eg using segmentation and/or tracking, but our local approach to motion estimation shows its limits in such extreme situations.

%====================================================================================================================
\subsection{Applicability to transparent motions and dynamic textures}

We tested the applicability of our method on scenes that are challenging (dynamic textures) or impossible (transparencies) to handle with traditional optical flow methods. There are no established benchmarks related to motion estimation and dynamic textures. Recent works \cite{derpanis2012,derpanis2010,teney2014,teney2015} that highlighted the potential of filter-based motion features in such situation focused on applications such as segmentation \cite{teney2014,teney2015} or scene recognition \cite{derpanis2012,derpanis2010}. In \fig\ref{fig:dynamictextures}, we show scenes containing dynamic textures (water, steam) from which we identified the dominant motion. The flow estimated by a typical method \cite{brox2004} is typically noisy and/or inaccurate, as the usual assumption of brightness constancy does not hold (\eg flickering effect on the water surface, changes of brightness/transparency of the steam, etc.). Although no ground truth is available for these scenes, the flow estimated by our methods is more reliable in comparison. We then demonstrate the ability of our distributed representation to capture multiple motions at a single location (transparencies and semi-transparencies), thus going beyond the optical flow representation of pixelwise displacements. We show features in \fig\ref{fig:dynamictextures}) from three sequences. The first depicts two alpha-blended (in equal proportions) textures moving in opposite directions, thus simulating transparency. The other two depict persons moving behind a fence in directions different than the fence itself \cite{derpanis2010}. Feature vectors from different locations in the image are visualized as radial bins (orientations) of concentric rings (speeds). Larger values (brighter bins) indicate motion evidence. As expected, areas with simple translations produce one major peak, whereas areas with transparencies produce correspondingly more complex, multimodal distributions\footnote{These experiments used a model trained on the Middlebury dataset.}.

\setlength{\fboxsep}{0pt}
\setlength{\fboxrule}{.3pt}

\begin{figure}[t]
  \begin{center}
  \scriptsize
  %\tiny
  %\vspace{-30pt}
  \renewcommand{\arraystretch}{0.8}
  \renewcommand{\tabcolsep}{0.40mm}
  \begin{tabular}{ccccl}
    \hangBox{\fbox{\includegraphics[height=11.5mm]{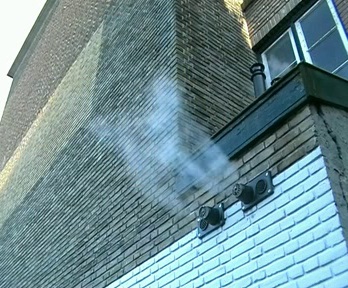}}}&
    \hangBox{\fbox{\includegraphics[height=11.5mm]{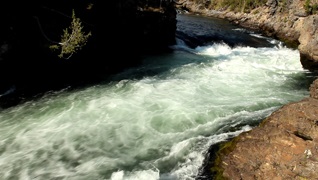}}}&
    \hangBox{\fbox{\includegraphics[height=11.5mm]{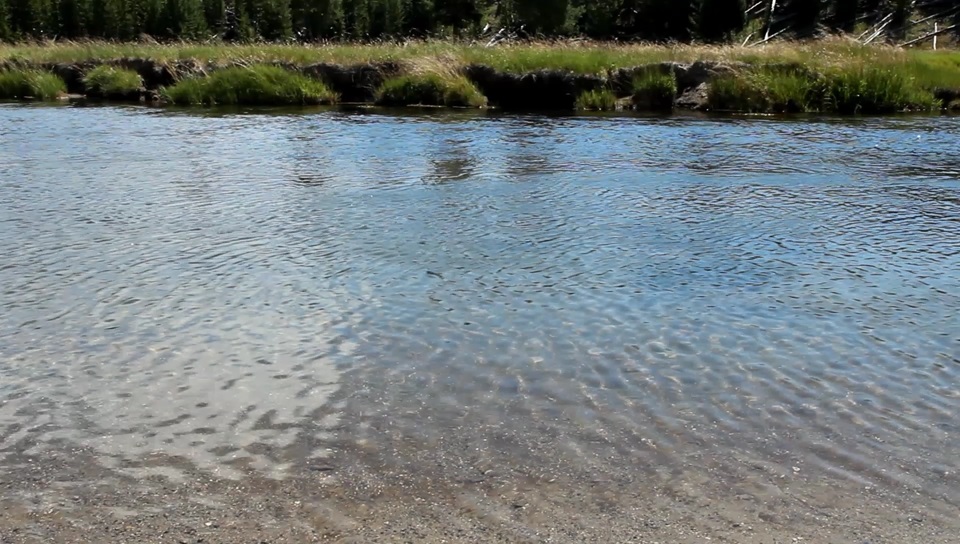}}}&
    \hangBox{\fbox{\includegraphics[height=11.5mm]{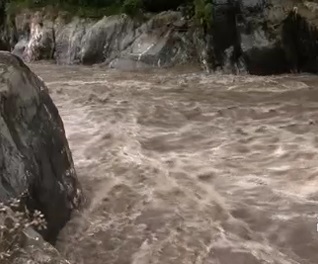}}}&
    \hangBox{\\~Input\\~frame}\\
    \hangBox{\fbox{\includegraphics[height=11.5mm]{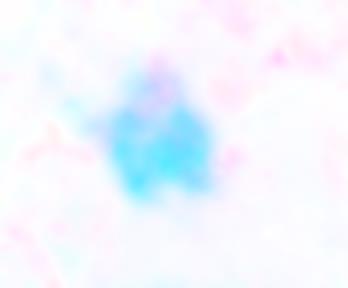}}}&
    \hangBox{\fbox{\includegraphics[height=11.5mm]{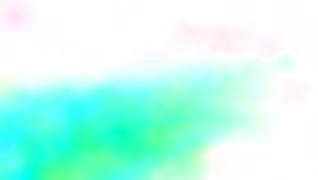}}}&
    \hangBox{\fbox{\includegraphics[height=11.5mm]{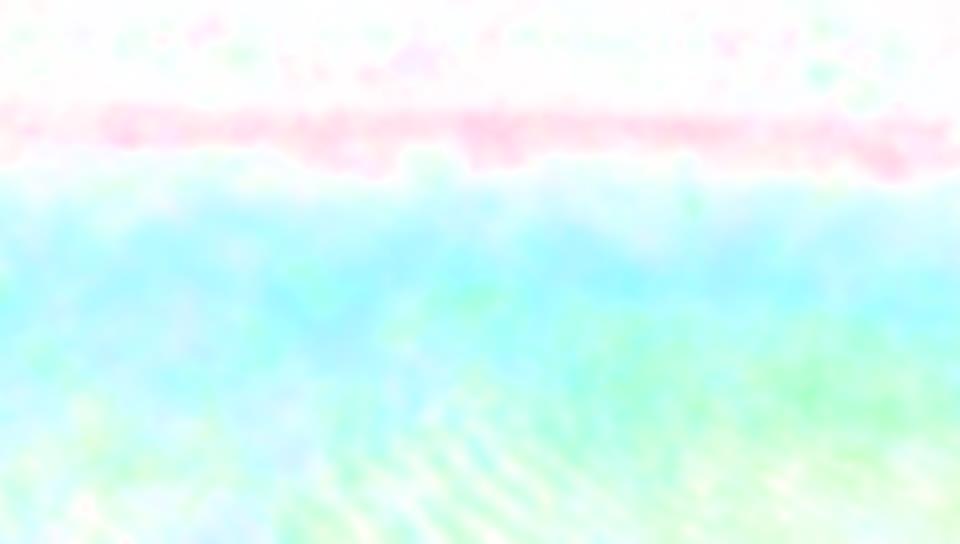}}}&
    \hangBox{\fbox{\includegraphics[height=11.5mm]{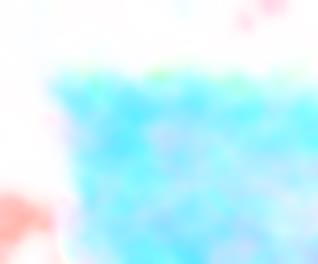}}}&
    \hangBox{\\~Flow as\\~proposed}\\
    \hangBox{\fbox{\includegraphics[height=11.5mm]{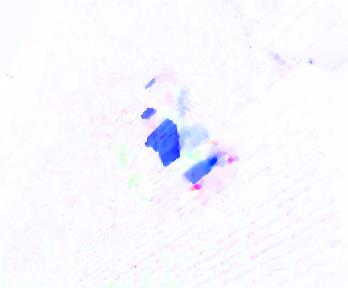}}}&
    \hangBox{\fbox{\includegraphics[height=11.5mm]{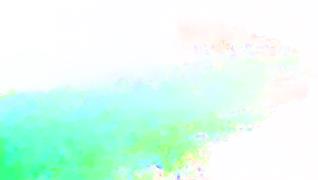}}}&
    \hangBox{\fbox{\includegraphics[height=11.5mm]{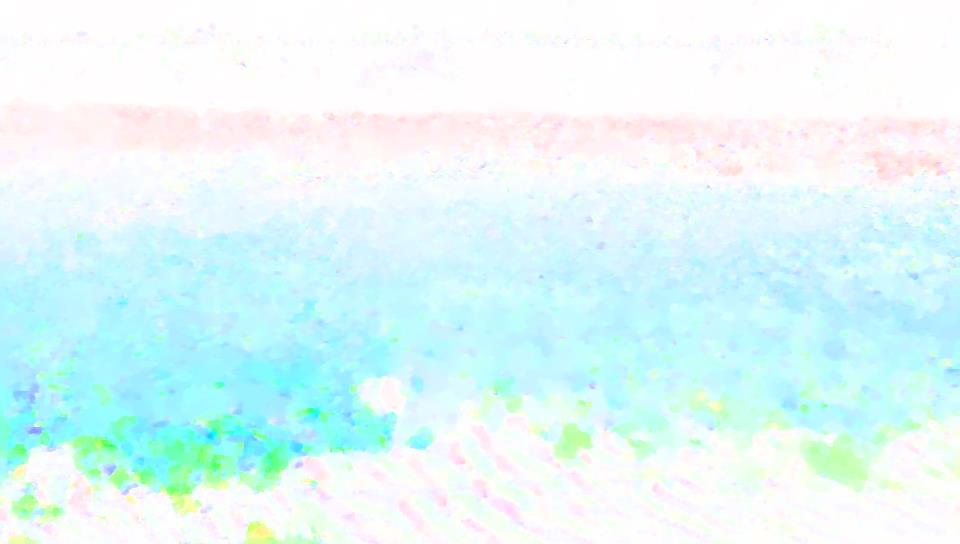}}}&
    \hangBox{\fbox{\includegraphics[height=11.5mm]{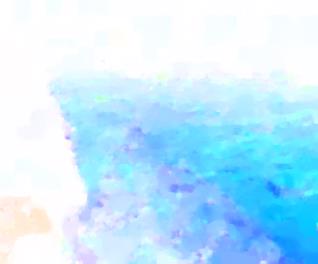}}}&
    \hangBox{\\~Flow\\~using \cite{brox2004}}\\
  \end{tabular}
  \end{center}
  %\vspace{-2pt}
  \caption{The extraction of optical flow on dynamic textures is challenging for traditional methods, as transparencies (\eg with steam, left) or flicker (\eg on water ripples, middle right) violate the typical assumption of brightness constancy. The core of our approach relies on the analysis of the frequency contents of the video, and produces more stable and reliable motion estimates.}
  \label{fig:dynamictextures}
  \vspace{-7pt}
\end{figure}

\begin{figure}[t]
  \begin{center}
  %\resizebox{.99\linewidth}{!}{
  \resizebox{\linewidth}{!}{
  \scriptsize
  %\vspace{-30pt}
  \renewcommand{\arraystretch}{1.45}
  \renewcommand{\tabcolsep}{1.2mm}
  \begin{tabular}{cccc}
    Input & Traditionnal & \multicolumn{2}{c}{{As proposed}} \vspace{-3pt} \\
    frame & optical flow \cite{brox2004} & \multicolumn{2}{c}{Dominant flow~/~Motion features} \\

    %\hangBox{\fbox{\includegraphics[width=21mm,height=19mm]{Reflections.png}}}&
    \hangBox{\fbox{\includegraphics[width=21mm,height=19mm]{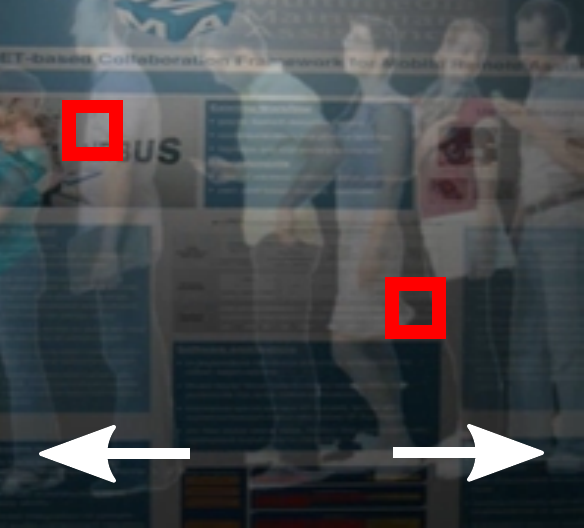}}}&
    \hangBox{\fbox{\includegraphics[width=21mm,height=19mm]{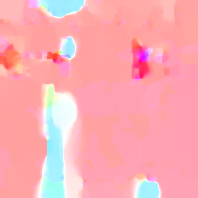}}}&
    \hangBox{\fbox{\includegraphics[width=21mm,height=19mm]{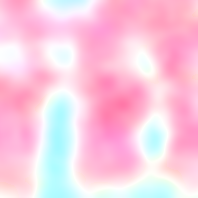}}}&
    \hangBox{\vspace{.2mm}\includegraphics[height=9mm]{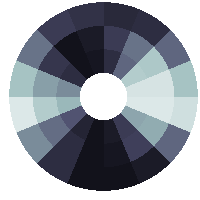}\\
    \includegraphics[height=9mm]{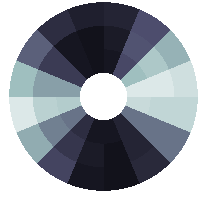}}\\

    %\hangBox{\fbox{\includegraphics[width=21mm,height=19mm]{Fence1.png}}}&
    \hangBox{\fbox{\includegraphics[width=21mm,height=19mm]{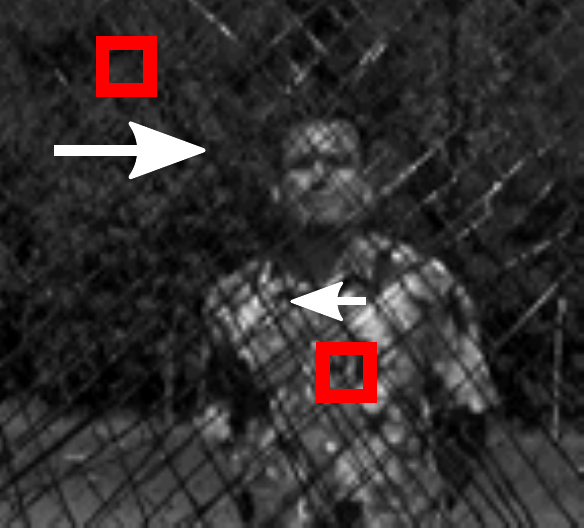}}}&
    \hangBox{\fbox{\includegraphics[width=21mm,height=19mm]{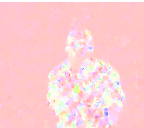}}}&
    \hangBox{\fbox{\includegraphics[width=21mm,height=19mm]{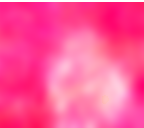}}}&
    \hangBox{\includegraphics[height=9mm]{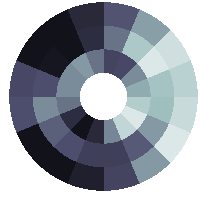}\\
    \includegraphics[height=9mm]{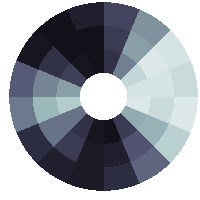}}\\

    %\hangBox{\fbox{\includegraphics[width=21mm,height=19mm]{Fence2.png}}}&
    \hangBox{\fbox{\includegraphics[width=21mm,height=19mm]{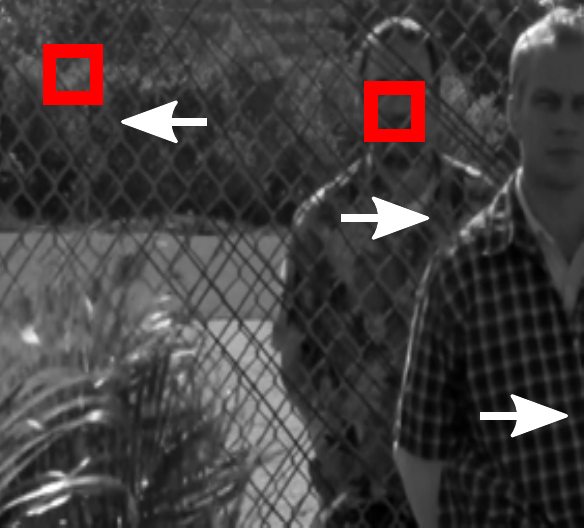}}}&
    \hangBox{\fbox{\includegraphics[width=21mm,height=19mm]{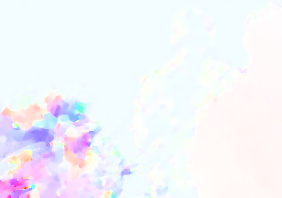}}}&
    \hangBox{\fbox{\includegraphics[width=21mm,height=19mm]{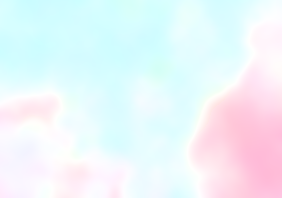}}}&
    \hangBox{\includegraphics[height=9mm]{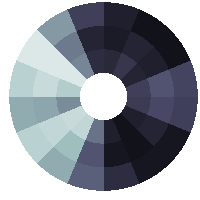}\\
    \includegraphics[height=9mm]{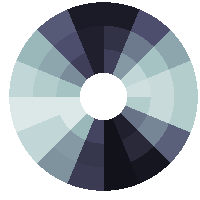}}\\
  \end{tabular}
  } % resizebox
  \end{center}
  %\vspace{-2pt}
  \caption{In scenes with multiple, transparent motions, traditional optical flow methods fail and typically produce incoherent results. Our method identifies a more stable dominant motion. More importantly, our higher-dimensional motion descriptor can capture multiple motions at single locations (red squares; white arrows indicate approximate ground truth direction of motion; see text for details).}
  \label{fig:dynamictextures}
  \vspace{-3pt}
\end{figure}

%====================================================================================================================
\section{Conclusions and future work}

We showed how to identify optical flow entirely within a convolutional neural network. By reasoning about required invariances and by using signal processing principles, we designed a simple architecture that can be trained end-to-end, from pixels to dense flow fields. We also showed how to enforce strict rotation invariance by constraining the weights, thus reducing the number of parameters and enabling training on small datasets. The resulting network performs on the Middlebury benchmark with performance comparable to classical methods, but inferior to the best engineered methods.

We believe the approach presented here bears two major advantages over existing optical flow algorithms. First, building upon the classical motion energy model, our approach is able to produce high-dimensional features that can capture non-rigid, transparent, or superimposed motions, which traditional optical flow cannot represent. Second, it constitutes a method for motion estimation formulated entirely as a shallow, easily-trainable CNN, without requiring any post-processing. Its potential is to be used as a building block in deeper architectures (\eg for activity or object recognition in videos) offering the possibility for fine-tuning the representation of motion. The potential of these two aspects will deserve further exploration and should be addressed in future work.

%====================================================================================================================

{\small
\bibliographystyle{ieee}
\bibliography{Bibliography}

\begin{thebibliography}{10}\itemsep=-1pt

\bibitem{adelson1985}
E.~H. Adelson and J.~Bergen.
\newblock Spatiotemporal energy models for the perception of motion.
\newblock {\em J. Opt. Soc. Am. A.}, 2, 284-299, 1985.

\bibitem{middlebury}
S.~Baker, D.~Scharstein, J.~Lewis, S.~Roth, M.~J. Black, and S.~R.
\newblock A database and evaluation methodology for optical flow.
\newblock In {\em International Conference on Computer Vision (ICCV)}, 2007.

\bibitem{brox2004}
T.~Brox, A.~Bruhn, N.~Papenberg, and W.~J.
\newblock High accuracy optical flow estimation based on a theory for warping.
\newblock In {\em European Conference on Computer Vision (ECCV)}, 2004.

\bibitem{brox2011}
T.~Brox and J.~Malik.
\newblock Large displacement optical flow: Descriptor matching in variational
  motion estimation.
\newblock {\em {IEEE} Transactions on Pattern Analysis and Machine Intelligence
  (PAMI)}, 33(3):500--513, 2011.

\bibitem{ldofflow}
T.~Brox and J.~Malik.
\newblock Large displacement optical flow: Descriptor matching in variational
  motion estimation.
\newblock {\em {IEEE} Transactions on Pattern Analysis and Machine Intelligence
  (PAMI)}, 2011.

\bibitem{sintel}
D.~J. Butler, J.~Wulff, G.~B. Stanley, and M.~J. Black.
\newblock A naturalistic open source movie for optical flow evaluation.
\newblock In {\em European Conference on Computer Vision (ECCV)}, 2012.

\bibitem{derpanis2010}
K.~G. Derpanis and R.~P. Wildes.
\newblock The structure of multiplicative motions in natural imagery.
\newblock {\em {IEEE} Transactions on Pattern Analysis and Machine Intelligence
  (PAMI)}, 32(7):1310--1316, 2010.

\bibitem{derpanis2012}
K.~G. Derpanis and R.~P. Wildes.
\newblock Spacetime texture representation and recognition based on a
  spatiotemporal orientation analysis.
\newblock {\em {IEEE} Transactions on Pattern Analysis and Machine Intelligence
  (PAMI)}, 34(6):1193--1205, 2012.

\bibitem{dieleman2015}
S.~Dieleman, K.~W. Willett, and J.~Dambre.
\newblock Rotation-invariant convolutional neural networks for galaxy
  morphology prediction.
\newblock {\em CoRR}, abs/1503.07077, 2015.

\bibitem{memin1998}
M.~E. and P.~P.
\newblock A multigrid approach for hierarchical motion estimation.
\newblock In {\em {IEEE Intenational Conference on Computer Vision (ICCV)}},
  1998.

\bibitem{fasel2006}
B.~Fasel and D.~Gatica-Perez.
\newblock Rotation-invariant neoperceptron.
\newblock In {\em International Conference on Pattern Recognition (ICPR)},
  2006.

\bibitem{flownet}
P.~Fischer, A.~Dosovitskiy, E.~Ilg, P.~H{\"{a}}usser, C.~Hazirbas, V.~Golkov,
  P.~van~der Smagt, D.~Cremers, and T.~Brox.
\newblock Flownet: Learning optical flow with convolutional networks.
\newblock {\em CoRR}, abs/1504.06852, 2015.

\bibitem{fleet1990}
D.~Fleet and A.~Jepson.
\newblock Computation of component image velocity from local phase information.
\newblock {\em International Journal of Computer Vision (IJCV)}, 5(1):77--104,
  1990.

\bibitem{fortun2015}
D.~Fortun, P.~Bouthemy, and C.~Kervrann.
\newblock Optical flow modeling and com putation: a survey.
\newblock {\em Computer Vision and Image Understanding (CVIU)}, 393, 2015.

\bibitem{heeger1987}
D.~J. Heeger.
\newblock Model for the extraction of image flow.
\newblock {\em J. Opt. Soc. Am. A}, 1987.

\bibitem{hornSchunk}
B.~Horn and S.~B.
\newblock Determining optical flow.
\newblock {\em Artificial Intelligence}, 11, 185-203, 1981.

\bibitem{jaderberg2015}
M.~Jaderberg, K.~Simonyan, A.~Zisserman, and K.~Kavukcuoglu.
\newblock Spatial transformer networks.
\newblock In {\em Advances in Neural Information Processing Systems (NIPS)},
  2015.

\bibitem{jayaraman2015}
D.~Jayaraman and K.~Grauman.
\newblock Learning image representations equivariant to ego-motion.
\newblock {\em CoRR}, abs/1505.02206, 2015.

\bibitem{karpathy2014}
A.~Karpathy, G.~Toderici, S.~Shetty, T.~Leung, R.~Sukthankar, and L.~Fei-Fei.
\newblock Large-scale video classification with convolutional neural networks.
\newblock In {\em {IEEE} Conference on Computer Vision and Pattern Recognition
  (CVPR)}, 2014.

\bibitem{konda2013}
K.~R. Konda and R.~Memisevic.
\newblock Unsupervised learning of depth and motion.
\newblock {\em CoRR}, abs/1312.3429, 2013.

\bibitem{krizhevsky2012}
A.~Krizhevsky, I.~Sutskever, and G.~E. Hinton.
\newblock Imagenet classification with deep convolutional neural networks.
\newblock In {\em Advances in Neural Information Processing Systems (NIPS)},
  2012.

\bibitem{laptev2015}
D.~Laptev and J.~M. Buhmann.
\newblock Transformation-invariant convolutional jungles.
\newblock In {\em {IEEE} Conference on Computer Vision and Pattern Recognition
  (CVPR)}, 2015.

\bibitem{le2010}
Q.~V. Le, J.~Ngiam, Z.~Chen, D.~Chia, P.~W. Koh, and A.~Y. Ng.
\newblock Tiled convolutional neural networks.
\newblock In {\em Advances in Neural Information Processing Systems (NIPS)},
  2010.

\bibitem{le2011}
Q.~V. Le, W.~Y. Zou, S.~Y. Yeung, and A.~Y. Ng.
\newblock Learning hierarchical invariant spatio-temporal features for action
  recognition with independent subspace analysis.
\newblock In {\em {IEEE} Conference on Computer Vision and Pattern Recognition
  (CVPR)}, 2011.

\bibitem{lecun1998b}
Y.~LeCun, L.~Bottou, G.~Orr, and K.~Muller.
\newblock Efficient backprop.
\newblock In G.~Orr and M.~K., editors, {\em Neural Networks: Tricks of the
  trade}. Springer, 1998.

\bibitem{niyogi1995}
S.~A. Niyogi.
\newblock Fitting models to distributed representations of vision.
\newblock In {\em International Joint Conference on Artificial Intelligence},
  pages 3--9, San Francisco, CA, USA, 1995. Morgan Kaufmann Publishers Inc.

\bibitem{olshausen2003}
B.~Olshausen.
\newblock Learning sparse, overcomplete representations of time-varying natural
  images.
\newblock In {\em ICIP}, volume~1, pages I--41--4 vol.1, 2003.

\bibitem{rowley1998}
H.~Rowley, S.~Baluja, and T.~Kanade.
\newblock Rotation invariant neural network-based face detection.
\newblock In {\em {IEEE} Conference on Computer Vision and Pattern Recognition
  (CVPR)}, 1998.

\bibitem{rust2006}
N.~C. Rust, V.~Mante, E.~P. Simoncelli, and J.~A. Movshon.
\newblock How mt cells analyze the motion of visual patterns.
\newblock {\em Nature Neuroscience}, 9, 2006.

\bibitem{simonyan2014}
K.~Simonyan and A.~Zisserman.
\newblock Two-stream convolutional networks for action recognition in videos.
\newblock {\em CoRR (NIPS Spotlight Session}, abs/1406.2199, 2014.

\bibitem{solari2015}
F.~Solari, M.~Chessa, N.~Medathati, and P.~Kornprobst.
\newblock What can we expect from a {V1-MT} feedforward architecture for
  optical flow estimation ?
\newblock {\em Signal Processing: Image Communication}, 2015.

\bibitem{sun2010}
D.~Sun, S.~Roth, and M.~J. Black.
\newblock Secrets of optical flow estimation and their principles.
\newblock In {\em {IEEE} Conference on Computer Vision and Pattern Recognition
  (CVPR)}, pages 2432--2439. IEEE, June 2010.

\bibitem{taylor2010}
G.~W. Taylor, R.~Fergus, Y.~LeCun, and C.~Bregler.
\newblock Convolutional learning of spatio-temporal features.
\newblock In {\em European Conference on Computer Vision (ECCV)}, 2010.

\bibitem{myWebsite0}
D.~Teney.
\newblock Personal website.
\newblock \url{http://damienteney.info/cnnFlow.htm}.

\bibitem{teney2014}
D.~Teney and M.~Brown.
\newblock Segmentation of dynamic scenes with distributions of spatiotemporally
  oriented energies.
\newblock In {\em British Machine Vision Conference (BMVC)}, 9 2014.

\bibitem{teney2015}
D.~Teney, M.~Brown, D.~Kit, and P.~Hall.
\newblock Learning similarity metrics for dynamic scene segmentation.
\newblock In {\em {IEEE} Conference on Computer Vision and Pattern Recognition
  (CVPR)}, 2015.

\bibitem{tran2014}
D.~Tran, L.~D. Bourdev, R.~Fergus, L.~Torresani, and M.~Paluri.
\newblock {C3D:} generic features for video analysis.
\newblock {\em CoRR}, abs/1412.0767, 2014.

\bibitem{ulman2010}
V.~Ulman.
\newblock Improving accuracy of optical flow of heeger's original method on
  biomedical images.
\newblock In {\em Image Analysis and Recognition}, volume 6111 of {\em Lecture
  Notes in Computer Science}, pages 263--273. Springer Berlin Heidelberg, 2010.

\bibitem{deepflow}
P.~Weinzaepfel, J.~Revaud, Z.~Harchaoui, and C.~Schmid.
\newblock {DeepFlow: Large displacement optical flow with deep matching}.
\newblock In {\em International Conference on Computer Vision (ICCV)}, 2013.

\bibitem{wu2015}
Z.~Wu, X.~Wang, Y.-G. Jiang, H.~Ye, and X.~Xue.
\newblock Modeling spatial-temporal clues in a hybrid deep learning framework
  for video classification.
\newblock In {\em ACM Multimedia Conference}, 2015.

\bibitem{ye2015}
H.~Ye, Z.~Wu, R.-W. Zhao, X.~Wang, Y.-G. Jiang, and X.~Xue.
\newblock Evaluating two-stream {CNN} for video classification.
\newblock In {\em ACM on International Conference on Multimedia Retrieval
  (ICMR)}, 2015.

\end{thebibliography}
}

\end{document}